\documentclass[fleqn,10pt]{wlscirep}
\usepackage[utf8]{inputenc}
\usepackage[T1]{fontenc}
\usepackage{float}
\usepackage{multirow}
\usepackage{diagbox}
\usepackage{makecell}
\usepackage{colortbl}
\usepackage{cleveref}

\title{DCNN: Dual Cross-current Neural Networks Realized Using An Interactive Deep Learning Discriminator for Fine-grained Objects}

\author[1,+]{Da Fu}
\author[2,*]{Mingfei Rong}
\author[2,*]{Eun-Hu Kim}
\author[3,+,*]{Hao Huang}
\author[4]{Witold Pedrycz}
\affil[1]{Shandong University, Department of Computer Science and Technology, Weihai, Shandong, 264200, China}

\affil[2]{Linyi University, Department of Big
Data, Linyi, Shandong, 276000, China}

\affil[3]{Suwon University, Department of Computer Science, Suwon, Hwaseong-si, Gyeonggi-do, 18323, South Korea}
\affil[4]{Alberta University, Department of Electrical \& Computer Engineering, Edmonton, T6G 2H5, Canada}

\affil[*]{corresponding author email: eunhu84@gmail.com; huanghao@suwon.ac.kr}

\affil[+]{these authors contributed equally to this work}

\begin{abstract}
    Accurate classification of fine-grained images remains a challenge in backbones based on convolutional operations or self-attention mechanisms. This study proposes novel dual-current neural networks (DCNN), which combine the advantages of convolutional operations and self-attention mechanisms to improve the accuracy of fine-grained image classification. The main novel design features for constructing a weakly supervised learning backbone model DCNN include (a) extracting heterogeneous data, (b) keeping the feature map resolution unchanged, (c) expanding the receptive field, and (d) fusing global representations and local features. Experimental results demonstrated that using DCNN as the backbone network for classifying certain fine-grained benchmark datasets achieved performance advantage improvements of 13.5--19.5\% and 2.2--12.9\%, respectively, compared to other advanced convolution or attention-based fine-grained backbones.
\end{abstract}
\begin{document}

\flushbottom
\maketitle
%
%
\thispagestyle{empty}


\section*{Introduction}

Convolutional neural networks~\cite{1,2,3,4} have been used to achieve considerable advances in computer vision tasks such as instance segmentation, image recognition, and object detection. Their success is mainly attributed to the convolutional operation, which learns local features hierarchically as powerful image representations. Despite its proficiency in extracting local features, it struggles to accurately discriminate fine-grained features owing to the limitation of a small receptive field, leading to poor accuracy in fine-grained image classification~\cite{41}. A feasible solution is to deepen the convolutional layers to expand the receptive field, such as in Very Deep Convolutional Networks (VGG)~\cite{6}, Residual Neural Networks (ResNet)~\cite{5}, and ConvNeXt~\cite{40}. However, they are still limited to learning local features while ignoring global representations, thus failing to improve the fine-grained image classification accuracy significantly.

\begin{figure}[htb]
  \centering
  \includegraphics[width=0.7\linewidth]{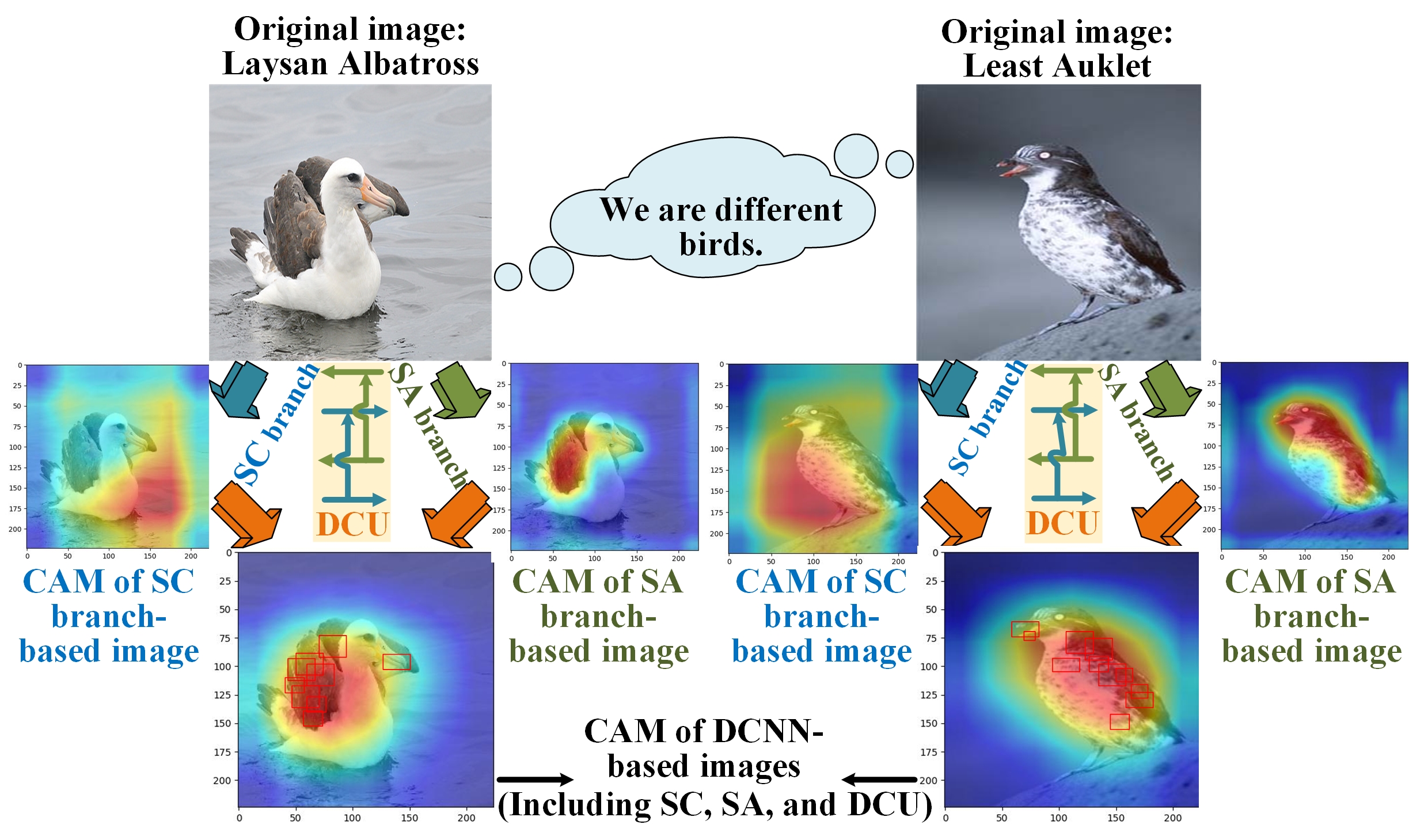}
  \caption{Class attention map (CAM) used for comparative evaluation of separable convolutional (SC) branch-based, self-attentional (SA) branch-based, and the proposed DCNN-based (including SC, SA, and dual cross-current unit (DCU)) bird image results on the CUB200-2011 dataset. The red-colored parts are the features that the model focuses on during feature extraction.}
  \label{fig:1}
\end{figure}

In recent times, self-attention mechanisms have been introduced in visual tasks~\cite{7,8,9,10}. The Vision Transformer (ViT)~\cite{11} and its related modification models, such as the Token-to-Token Vision Transformer (T2T-ViT)~\cite{12}, the data-efficient image transformers (DeiT)~\cite{13}, and the Swin Transformer~\cite{39} segment each image into patches with positional embeddings to construct a sequence of markers. It uses multi-head self-attention mechanisms to learn features to use as visual representations. Owing to the multilayer perceptron (MLP) and self-attention mechanisms, vision transformers reflect feature dependencies and spatial transformations over long distances, forming global representations. Unfortunately, existing self-attention mechanism-based networks ignore local features, thus reducing the identifiability between foreground and background~\cite{42}.

Self-attention mechanisms capable of acquiring global representations still do not significantly improve the accuracy of fine-grained image recognition tasks because they neglect local features. It is still unclear how accurately fine-grained features can be discriminated.

To improve the classification accuracy of fine-grained images, DCNN is proposed. Its structure design is shown in~\cref{fig:1}. The input image is heterogeneously composed into feature matrices and feature vectors, which are subjected to the separable convolutional (SC) branch (including depth-wise/point-wise convolutions~\cite{14}, gaussian error linear units (GELU)~\cite{38} activation function), and BatchNorm (BN)~\cite{15}) for local feature extraction and the self-attentional (SA) branch (including multi-head self-attention, MLP, and LayerNorm~\cite{16}) for global representations learning, respectively. The DCU continuously removes the differences in heterogeneous data~\cite{17} and branch semantics between the SA and SC branches and enhances the fusion of local features and global representations. The visualization of the Class Attention Map (CAM)~\cite{18} shows that, unlike ResNet and ViT, DCNN aims to couple local features based on large receptive field convolution operations with self-attention-based global representations to improve the accuracy of classifying fine-grained images by enhancing fine-grained feature learning.

Considering the mismatch between SC and SA branches, an interactive learning bridge, DCU, is designed. To fuse the two types of heterogeneous features, the DCU utilizes Up and Down strategies to match the feature resolution of two branches, 1~$\times$~1 convolution to align the channel size, BatchNorm, and LayerNorm to align the feature values, and concatenation to accumulate feature information from all previous bridges into the current bridge to enhance the interactivity of the two branches and improve classification performance. Furthermore, as the SC and SA branches prefer to capture features from different perspectives, DCUs are inserted into each block to interactively and successively remove semantic divergences between them. This fusion procedure substantially enhances the local details of the global representations and the global perception of local features.

The contributions of this study are as follows:
\begin{itemize}
  \item We propose a novel DCNN in which SC and SA branches handle heterogeneous data. While keeping the feature block resolution unchanged, the DCNN maximizes the preservation of global representations and local features of the original data, thus improving the classification accuracy of fine-grained images;
  \item We propose DCU, which generates interactive learning bridges between the SC and SA branches to fuse the local features with the global representations. To improve the interactive learning ability of the SC and SA branches, the historical feature information of all the previous bridges is accumulated into the current bridge through concatenation (Concat);
  \item DCNN was used to perform classification on ImageNet-1k, fine-grained benchmark datasets, and the medical brain tumor MRI dataset. DCNN has fewer parameters and higher classification accuracy than other advanced methods.
\end{itemize}

The paper is organized as follows: some related works are reviewed in Section 2. Section 3 introduces the structure of dual cross-current neural networks. Section 4 evaluates the experiments. The conclusion is given in Section 5.

\section*{Related works}

\textbf{Convolution operation}. Deep learning networks based on convolution operations have strong local feature extraction capabilities in image blocks having diverse resolutions but insufficient ability to obtain global representations. Obtaining only local features is not enough to effectively distinguish fine-grained features. To alleviate this limitation, Zheng et al.~\cite{19} proposed the Trilinear Attention Sampling Network (TASN), which learns ﬁne-grained details through a trilinear attention module and models the relationships between channels to generate attention maps. An attention-based sampler takes the attention maps and images as input, highlighting the attended parts in high resolution. The final classification was performed using the convolution-based backbone models. Ji et al.~\cite{20} proposed the Attention Convolutional Binary Neural Tree (ACNet), which incorporates convolutional operations at the edges of the tree structure. It determines the computational paths from the root to the leaves within the tree by using a routing function at each node. The final decision is computed by summing the predictions from the leaf nodes. The convolution-based backbone models were used to classify the preprocessed feature maps.

The classification backbone model used in the aforementioned methods is based on convolutional operations and is limited to local feature learning. Our proposed backbone model, DCNN, learns local features and acquires global representations, improving fine-grained feature recognition.

\textbf{Attention mechanism}. In computer vision, attention mechanisms are generally classified into channel, spatial, and hybrid domains. The principle is to assign different weights to channels or domains instead of treating a location or all channels in space as having the same weight when performing convolution/pooling operations as in the past, thus allowing the network to focus on extracting more critical information~\cite{21,22,23}. This attention mechanism of assigning different weights ignores important local feature information, making the model more inclined towards a global representation. However, identifying and learning local features in fine-grained image regions is required to distinguish fine-grained image categories better. Wang et al.~\cite{24} presented the Feature Fusion Vision Transformer (FFVT), which automatically detects distinguished regions and exploits global and local information at multiple levels in the image through mutual attention weight selection, enabling efficient selection of the information markers that are highly similar to the class. Context the information tokens that have high similarity with the class tokens. Finally, the attention-based backbone models were used to classify images with well-labeled features. Hu et al.~\cite{25} adaptively selected the most discriminative regions for each image using the proposed dynamic patch proposal module (DPPM). They built a finer-scale network to take the trend parts upscaled by previous scales as input cyclically, which enters the attention-based backbone networks, which perform the fine-grain map classification. He et al.~\cite{26} presented TransFG, which identifies discriminative regions and removes redundant information through the Part Selection Module. Furthermore, contrast loss was introduced to improve the discriminability, and attention-based backbone models were used to classify fine-grained images.

The classification backbone model used in the aforementioned methods relies on attention mechanisms; it is better at global feature extraction but ignores local feature learning. Our proposed backbone model, DCNN, can acquire global representations and learn local features to improve fine-grained feature recognition.

\section*{Methods}
\subsection*{Overview}

\begin{figure*}[t]
  \centering
  \includegraphics[width=1.0\linewidth]{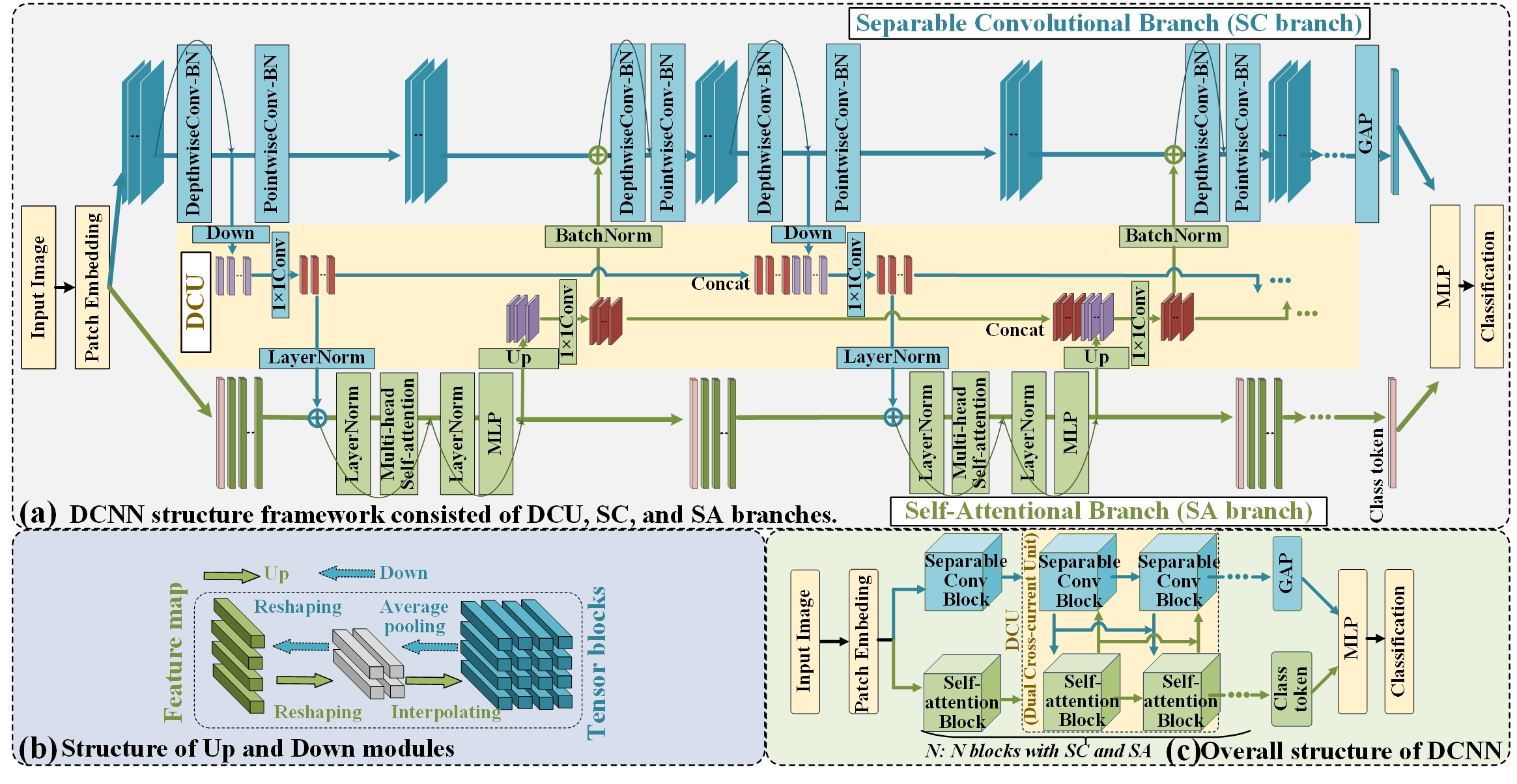}
  \caption{Overall DCNN network framework and its inherent design methodologies. (a) Detailed construction process of SC branch, DCU, and SA branch. (b) Up and Down modules for spatial alignment of patch embeddings and feature maps. (c) Thumbnail sketch of the DCNN. Here, BN denotes BatchNorm, GAP denotes Global Average Pooling.}
  \label{fig:2}
\end{figure*}

To enhance the differentiation of fine-grained features using local features as well as global representations, we designed the DCNN, as shown in~\cref{fig:3}. The SC branch collects local features hierarchically through a separable convolution operation. It retains the local features as feature blocks. The local features reflect a compact feature representation of the local matrix neighborhood of the fine-grained image. The SA branch aggregates the global representations in compressed patch embedding vectors through self-attention modules. Global representations reflect the long-range feature dependencies of embedding vectors. The DCU bridges continuously generate interactions between SC-branch feature blocks and SA-branch embedding vectors to fuse global representations and local features to enhance fine-grained feature discriminability.

\definecolor{myblue}{RGB}{189,214,238}
\definecolor{myyellow}{RGB}{255,242,204}
\definecolor{mygreen}{RGB}{226,239,217}
\begin{table*}
  \centering
  \resizebox{1.0\textwidth}{!}{
    \begin{tabular}{c|c|cc|cc|cc|c}
      \specialrule{0.08em}{0pt}{0pt}
      Stage                    & S1             & \multicolumn{2}{c|}{S2-S5}                       & \multicolumn{2}{c|}{S6-S9}          & \multicolumn{2}{c|}{S10-S13}        & S14                                                                                                       \\
      \hline
      Output                   & 56~$\times$~56 & \multicolumn{2}{c|}{56~$\times$~56}              & \multicolumn{2}{c|}{56~$\times$~56} & \multicolumn{2}{c|}{56~$\times$~56} & 1~$\times$~1                                                                                              \\
      \hline
      \cellcolor{myblue}       &                & \cellcolor{myblue}[1~$\times$~1, 32, stride 1]   & \multirow{4}{*}{$\times$~1}         & \cellcolor{myblue}[1~$\times$~1, 64, stride 1]        & \multirow{4}{*}{$\times$~1} & \cellcolor{myblue}[1~$\times$~1, 128, stride 1] & \multirow{4}{*}{$\times$~1} &               \\
      \cellcolor{myblue}       &                & \cellcolor{myblue}[9~$\times$~9, 32, stride 1]   &                                     & \cellcolor{myblue}[9~$\times$~9, 64, stride 1]        &                             & \cellcolor{myblue}[9~$\times$~9, 128, stride 1] &                             &               \\
      \cellcolor{myblue}       &                & \cellcolor{myblue} [1~$\times$~1, 32, stride 1]  &                                     & \cellcolor{myblue}[1~$\times$~1, 64, stride 1]        &                             & \cellcolor{myblue}[1~$\times$~1, 128, stride 1] &                             &               \\
      \cellcolor{myblue}SC     &                & \cellcolor{myblue} [1~$\times$~1, 128, stride 1] &                                     & \cellcolor{myblue}[1~$\times$~1, 256, stride 1]       &                             & \cellcolor{myblue}[1~$\times$~1, 512, stride 1] &                             & 1~$\times$~1, \\
      \cline{3-8}
      \cellcolor{myblue}branch &                & \cellcolor{myblue}[1~$\times$~1, 32, stride 1]                     & \multirow{4}{*}{$\times$~3}         & \cellcolor{myblue}[1~$\times$~1, 64, stride 1]        & \multirow{4}{*}{$\times$~3} & \cellcolor{myblue}[1~$\times$~1, 128, stride 1] & \multirow{4}{*}{$\times$~3} & 1000          \\
      \cellcolor{myblue}       &                & \cellcolor{myblue}[9~$\times$~9, 32, stride 1]                     &                                     & \cellcolor{myblue}[9~$\times$~9, 64, stride 1]        &                             & \cellcolor{myblue}[9~$\times$~9, 128, stride 1] &                             &               \\
      \cellcolor{myblue}       & 4~$\times$~4,  & \cellcolor{myblue}[1~$\times$~1, 32, stride 1]                     &                                     & \cellcolor{myblue}[1~$\times$~1, 64, stride 1]        &                             & \cellcolor{myblue}[1~$\times$~1, 128, stride 1] &                             &               \\
      \cellcolor{myblue}       & 64 stride 4    & \cellcolor{myblue}[1~$\times$~1, 256, stride 1]                    &                                     & \cellcolor{myblue}[1~$\times$~1, 256, stride 1]       &                             & \cellcolor{myblue}[1~$\times$~1, 512, stride 1] &                             &               \\
        \cmidrule{1-1}\cmidrule{3-9}
        \cellcolor{myyellow} & &
        \multicolumn{2}{c|}{\begin{tabular}{@{}cccc@{}}\cellcolor{myyellow}[1~$\times$~1, 576]& \phantom{$\downarrow$}&\cellcolor{myyellow}[1~$\times$~1, 32]&\phantom{$\uparrow$}\\\end{tabular}}&
        \multicolumn{2}{c|}{\begin{tabular}{@{}cccc@{}}\cellcolor{myyellow}[1~$\times$~1, 576]& \phantom{$\downarrow$}&\cellcolor{myyellow}[1~$\times$~1, 64]&\phantom{$\uparrow$}\\\end{tabular}}&
        \multicolumn{2}{c|}{\begin{tabular}{@{}cccc@{}}\cellcolor{myyellow}[1~$\times$~1, 576]& \phantom{$\downarrow$}&\cellcolor{myyellow}[1~$\times$~1, 128]&\phantom{$\uparrow$}\\\end{tabular}} &\multirow{4}{*}{-}\\
        \cellcolor{myyellow}& &
        \multicolumn{2}{c|}{\begin{tabular}{@{}cccc@{}}\cellcolor{myyellow}[1~$\times$~1, 576]& \phantom{$\downarrow$}&\cellcolor{myyellow}[1~$\times$~1, 32]&\phantom{$\uparrow$}\\\end{tabular}}&
        \multicolumn{2}{c|}{\begin{tabular}{@{}cccc@{}}\cellcolor{myyellow}[1~$\times$~1, 576]& \phantom{$\downarrow$}&\cellcolor{myyellow}[1~$\times$~1, 64]&\phantom{$\uparrow$}\\\end{tabular}}&
        \multicolumn{2}{c|}{\begin{tabular}{@{}cccc@{}}\cellcolor{myyellow}[1~$\times$~1, 576]& \phantom{$\downarrow$}&\cellcolor{myyellow}[1~$\times$~1, 128]&\phantom{$\uparrow$}\\\end{tabular}} & \\
        \cellcolor{myyellow} & &
        \multicolumn{2}{c|}{\begin{tabular}{@{}cccc@{}}\cellcolor{myyellow}[1~$\times$~1, 576]& $\downarrow$&\cellcolor{myyellow}[1~$\times$~1, 32]&$\uparrow$\\\end{tabular}}&
        \multicolumn{2}{c|}{\begin{tabular}{@{}cccc@{}}\cellcolor{myyellow}[1~$\times$~1, 576]& $\downarrow$&\cellcolor{myyellow}[1~$\times$~1, 64]&$\uparrow$\\\end{tabular}}&
        \multicolumn{2}{c|}{\begin{tabular}{@{}cccc@{}}\cellcolor{myyellow}[1~$\times$~1, 576]& $\downarrow$&\cellcolor{myyellow}[1~$\times$~1, 128]&$\uparrow$\\\end{tabular}} & \\
        \cellcolor{myyellow}\multirow{-4}{*}{DCU}& &
        \multicolumn{2}{c|}{\begin{tabular}{@{}cccc@{}}\cellcolor{myyellow}[1~$\times$~1, 576]& \phantom{$\downarrow$}&\cellcolor{myyellow}[1~$\times$~1, 32]&\phantom{$\uparrow$}\\\end{tabular}}&
        \multicolumn{2}{c|}{\begin{tabular}{@{}cccc@{}}\cellcolor{myyellow}[1~$\times$~1, 576]& \phantom{$\downarrow$}&\cellcolor{myyellow}[1~$\times$~1, 64]&\phantom{$\uparrow$}\\\end{tabular}}&
        \multicolumn{2}{c|}{\begin{tabular}{@{}cccc@{}}\cellcolor{myyellow}[1~$\times$~1, 576]& \phantom{$\downarrow$}&\cellcolor{myyellow}[1~$\times$~1, 128]&\phantom{$\uparrow$}\\\end{tabular}} & \\
        \cmidrule{1-1}\cmidrule{3-9}
        \cellcolor{mygreen} &                           & \cellcolor{mygreen}[MHSA (12), 576]        &          & \cellcolor{mygreen}[MHSA (12), 576]        &  &\cellcolor{mygreen} [MHSA (12), 576] &  & \multirow{3}{*}{\makecell[c]{1~$\times$~1,\\1000}} \\
        \cellcolor{mygreen}&                           & \cellcolor{mygreen}Linear (576, 2304)        &          $\times$~3                          & \cellcolor{mygreen}Linear (576, 2304)       &            $\times$~3                 & \cellcolor{mygreen}Linear (576, 2304) &         $\times$~3                    &  \\
       \cellcolor{mygreen}\multirow{-3}{*}{\makecell[c]{SA\\branch}} &                         & \cellcolor{mygreen}Linear (2304, 576)        &                                     & \cellcolor{mygreen}Linear (2304, 576)        &                             & \cellcolor{mygreen}Linear (2304, 576) &                             &  \\
        \specialrule{0.08em}{0pt}{0pt}
    \end{tabular}}
  \caption{Detailed settings of the DCNN structure. 56~$\times$~56 indicates the size of the feature map. MHSA (12) means 12 heads in the SA branch with multi-head self-attention. ``$\downarrow$'' denotes Down module, ``$\uparrow$'' denotes Up module.}
  \label{table:1}
\end{table*}

\subsection*{Structure analysis of Dual Cross-current Neural Networks}

\noindent\textbf{SC branch}. The SC branch uses the strategy of keeping the feature map resolution unchanged. As shown in~\cref{fig:3} (a), the entire branch is divided into three stages, as described in~\cref{table:1} (SC branch). Each stage consists of four convolution blocks, each containing four bottlenecks, each with 1~$\times$~1 down-projected convolution, 9~$\times$~9 depth-wise convolution, 1~$\times$~1 point-wise convolution, 1~$\times$~1 up-projected convolution, and the residual connections between the inputs and outputs of the bottlenecks.

Convolution operation with the number of input channels $c_{in}$, the number of output channels $c_{out}$, and a convolution kernel of $p$ size and a step size of $p$, which can achieve patch embedding.
\begin{equation}
z_0^{SC} = BN\{\sigma[Con{v_{{c_{in}} \to {c_{out}}}}(X,s = k\_size = p)]\}
\end{equation}

The SC branch consists of depth-wise and point-wise convolutions, as shown in~\cref{fig:4}. The activation function and BatchNorm are performed after each convolution. The depth-wise convolution can be written as follows:
\begin{equation}
z_L^{SC} = BN\{\sigma[DepthwiseConv(z_{L - 1}^{SC})]\}  + z_{L - 1}^{SC}
\end{equation}
The point-wise convolution can be written as follows:
\begin{equation}
z_{L + 1}^{SC} = BN\{ \sigma[PointwiseConv(z_L^{SC})]\}
\end{equation}
where $BN$ means BatchNorm, $X$ is the image input, $s$ denotes stride, $k\_size$ denotes kernel size, $L\in N^+$, and $\sigma$ denotes GELU activation function can be written as:
\begin{equation}
\sigma(x) = 0.5x \times \{ 1 + \tanh [\sqrt {\pi / 2}
(x + 0.047715{x^3})]\}
\end{equation}

\noindent\textbf{SA branch}. The SA branch contains N repeating transformer blocks from T2T-ViT. Each transformer block consists of a multi-headed self-attention module and an MLP module, as shown in~\cref{fig:3} (a). LayerNorms were applied before the remaining connection of the self-attention layer and the MLP module. We use 4~$\times$~4 convolution with step size 4 to compress the generated feature maps into 14~$\times$~14 non-overlapping patch embeddings and pass them through a linear projection layer. Classification is then performed.

\noindent\textbf{DCU}. We design DCU to continuously remove the heterogeneous data differences between the feature map matrix of the SC branch and the patch embedding vectors of the SA branch in an extended interactive manner. The SC feature map block is a tensor composed of channels, heights, and widths, while the patch embedding is a matrix consisting of the number of embedding dimensions and image patches.

When feeding back from the SC branch to the SA branch, the spatial dimension alignment is first performed using the Down module (\cref{fig:3}(b)). Then, the feature map must be aligned to the number of channels of the patch embedding by 1~$\times$~1 convolution. Finally, the feature map is added to the patch embedding as shown in~\cref{fig:3}(a).
\begin{equation}
b_L^{SC} = Conv1[Down(z_L^{SC}) + b_{L - 1}^{SC}]
\end{equation}
\begin{equation}
z_{L + 1}^{SA} = LN\{ Conv1[Down(z_L^{SC}) + b_L^{SC}]\}
\end{equation}

When feeding back from the SA to the SC branches, the patch embedding needs an Up module to adjust the spatial scale (\cref{fig:3}(b)). It is adjusted to the dimensions of the SC branch channels by 1~$\times$~1 convolution and added to the feature map.
\begin{equation}
b_L^{SA} = Conv1[Up(z_L^{SA}) + b_{L - 1}^{SA}]
\end{equation}
\begin{equation}
z_{L + 1}^{SC} = BN\{ Conv1[Up(z_L^{SA}) + b_L^{SA}]\}
\end{equation}
where $BN$ means BatchNorm, $LN$ means LayerNorm, $b^{SC}$ is the DCU bridge from SC branch to SA branch, $b^{SA}$ is the DCU bridge from SA branch to SC branch, and $L\in N^+$. The detailed settings of the model convolution operation are listed in \cref{table:1}.

\section*{Results and discussion}

\subsection*{Experimental setups}

This section describes the experimental setups. DCNN was implemented using PyTorch. All the results were evaluated on a system with an AMD Ryzen93950X 16-Core Processor 3.49 GHz CPU, NVIDIA GeForce RTX 4090 GPU with 24 GB of memory, running on Windows 10 (64-bit). To evaluate the performance of the DCNN against other baseline models, all models were trained using the Adamw optimizer to update weights. The cross-entropy loss function~\cite{27} was applied to the DCNN for the ImageNet1k dataset, and the center loss function~\cite{28} was used in the DCNN for fine-grained datasets.\cref{table:2} lists six experimental datasets used, including ImageNet-1k, four fine-grained datasets (without box-bounding markers), and one applied brain tumor MRI dataset.

\begin{table}[H]
    \centering
        \begin{tabular}{cccc}
            \toprule
            Dataset name       & Meta-class & \#Images & \#Categories \\
            \midrule
            ImageNet-1k~\cite{29}   & -          & 14197122 & 1000         \\
            Oxford Flowers~\cite{30}  & Flowers    & 8189     & 102          \\
            Food101~\cite{31}       & Foods      & 101000   & 101          \\
            Stanford Dogs~\cite{32} & Dogs       & 48562    & 555          \\
            CUB200-2021~\cite{33}   & Birds      & 11788    & 200          \\
            Brain tumor MRI    & Tumors     & 10183    & 4            \\
            \bottomrule
        \end{tabular}
    \caption{Detailed statistics of datasets used in this study.}
    \label{table:2}
\end{table}

\subsection*{Experimental analysis in ImageNet-1k dataset}
The experiments are evaluated on the ImageNet 2012 classification dataset containing 1000 categories. Our proposed DCNN and these advanced models for comparison were trained on 1.28 million training images and evaluated on 50000 testing images.
\begin{table}[htb]
    \centering
        \begin{tabular}{cccc}
            \toprule
            Models              & \#Params  & \#FLOPs  & Top1-acc \\
            \midrule
            ResNet-152~\cite{5}      & 60.20M  & 11.6G  &79.0\%   \\
            ViT-B~\cite{11}          & 86.12M  & 55.5G  & 77.9\%   \\
            ViT-L~\cite{11}          & 306.96M & 191.1G & 76.5\%   \\
            T2T-ViT-19~\cite{12}     & 64.15M  & 13.2G  & 81.2\%   \\
            DeiT-B~\cite{13}         & 86.01M  & 17.6G  & 81.6\%   \\
            MLP-Mixer~\cite{34}      & 59.46M  & 11.4G  & 76.3\%   \\
            ResMLP-B~\cite{35}       & 129.00M  & 27.1G  & 80.7\%   \\
            ConvMixer-1536/20~\cite{37} & 51.62M  & 11.0G  & 81.2\%   \\
            RegNetY-32.0GF~\cite{36} & 145.08M & 32.3G  &80.4\%   \\
            SwinT-B~\cite{39}     & 88.00M  & 15.4G  & 83.5\%   \\
            ConvNeXt-L(iso.)~\cite{40}     & 87.00M  & 16.9G  & 82.0\%   \\
            ConvNeXt-B~\cite{40}     & 89.00M  & 15.4G  & 83.8\%   \\
            DCNN (Ours)     & 52.68M  & 11.0G  & \textbf{82.3}\%  \\
            DCNN-B (Ours)     & 88.47M  & 16.1G  & \textbf{84.0\%}   \\
            \bottomrule
        \end{tabular}
    \caption{Comparison of classification performance between the proposed DCNN and other advanced models. We evaluate Params (number of parameters), FLOPs (floating point operations), and Top1-acc (top-1 accuracy) on the ImageNet-1k dataset. Adam optimized all models with a batch size of 128, a learning rate of 0.001, and a training process of 300 epochs. DCNN-B is a larger model that builds on the DCNN architecture by increasing the number of depth-wise/point-wise convolutions and self-attention heads.}
    \label{table:3}
\end{table}

With similar computational budgets, the DCNN outperforms the convolution operation-based or attention
mechanism-based models, as seen in \cref{table:3}. For example, DCNN (with 52.68M parameters and 11.0G FLOPs) has 12.5\% (52.68M vs. 60.20M) fewer parameters
than ResNet-152 (with 60.20M parameters and 11.6G FLOPs), 5.2\% (11.0G vs. 11.6G) less FLOPs cost, and a
performance improvement of 4.2\% (82.3\% vs. 79.0\%); DCNN has 38.8\% fewer parameters (52.68M vs. 86.12M), 80.2\% lower FLOPs cost (11.0G vs. 55.5G), and 5.6\% higher performance (82.3\% vs. 77.9\%) than ViT-B (which has 86.12M parameters and 55.5G of FLOPs); DCNN has 82.8\% (52.68M vs. 306.96M) fewer parameters than ViT-L (which has 306.96M parameters and 191.1G FLOPs), 94.2\% (11.0G vs. 191.1G) lower FLOPs cost, and 7.6\% (82.3\% vs. 76.5\%) performance improvement.

Our proposed DCNN, with fewer parameters and moderate FLOPs cost, outperforms other advanced models such as T2T-ViT, DeiT, MLP-Mixer, ResMLP, and RegNetY, demonstrating its superior performance as a backbone network. At comparable parameter scales, the performance of our proposed DCNN is also competitive with state-of-the-art ConvMixer, Swin Transformer (SwinT), and ConvNeXt.

\subsection*{Experimental analysis in fine-grained benchmarks}

\begin{figure*}[t]
  \centering
  \includegraphics[width=0.8\linewidth]{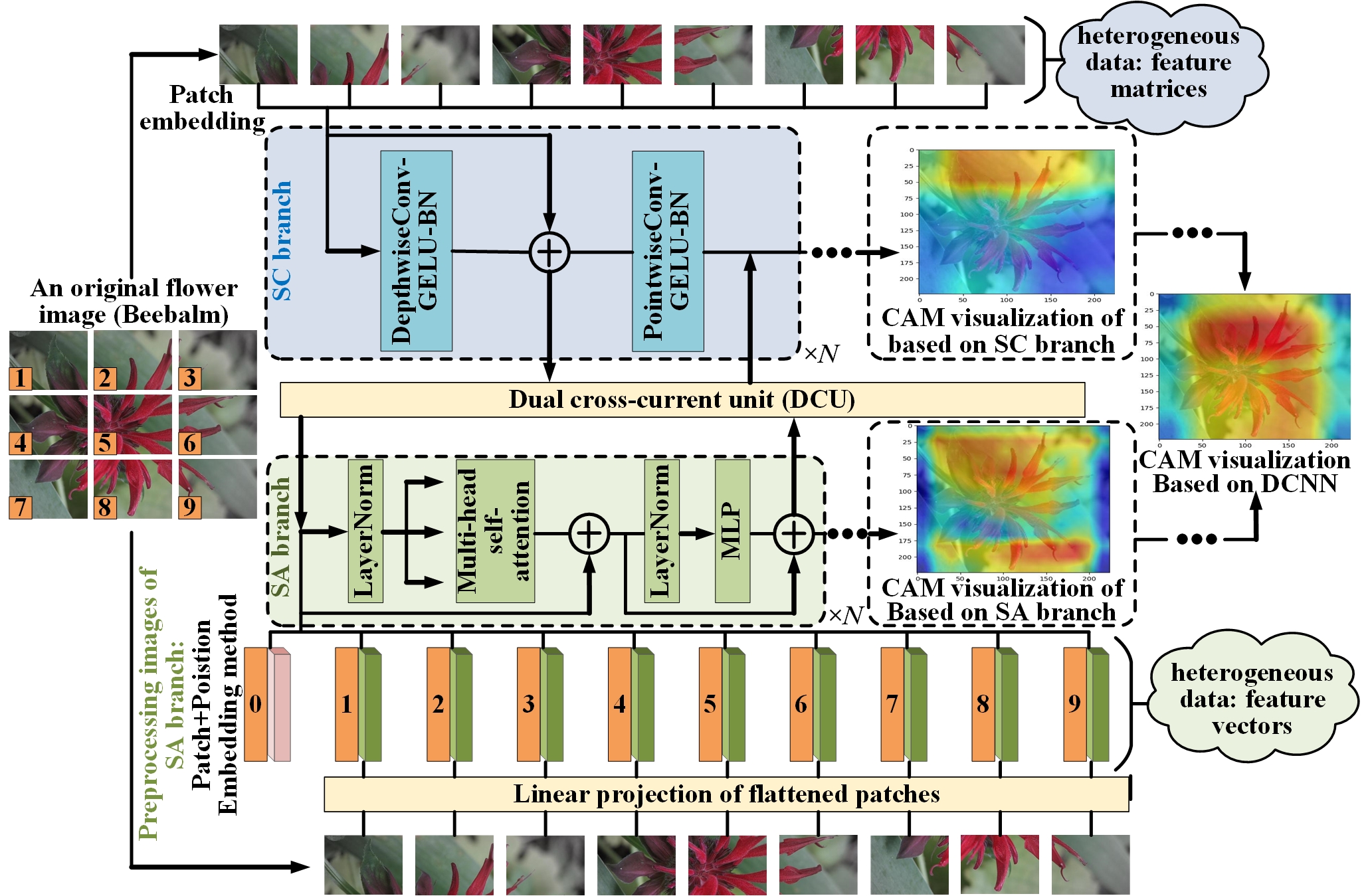}
  \caption{Comparison of the visualization of CAM between the proposed DCU-base and SC-base/SA-base demonstrated through the learning process of the DCNN on a flower image input. The red-colored parts are the features that the model focuses on during feature extraction.}
  \label{fig:2}
\end{figure*}

\definecolor{mygrey}{RGB}{220,220,220}
\begin{table*}[!htb]
    \centering
    \resizebox{0.9\linewidth}{!}{
        \begin{tabular}{cccccccccc}
            \toprule
                                               & \multirow{3}{*}{\diagbox{\rule[-2mm]{0pt}{2mm}Pre-Methods}{\rule{0pt}{4mm}Datasets}} &
            \multicolumn{2}{c}{Oxford Flowers} &
            \multicolumn{2}{c}{Food101}        &
            \multicolumn{2}{c}{Stanford Dogs}  &
            \multicolumn{2}{c}{CUB200-2021}                                                                                                                                                                           \\
            \cmidrule(r){3-4}\cmidrule(lr){5-6}\cmidrule(lr){7-8}\cmidrule(l){9-10}
            Mechanisms                          &                                                                                      & Top1-   & Top5-   & Top1-   & Top5-   & Top1-   & Top5-   & Top1-   & Top5-   \\
                                               &                                                                                      & acc(\%) & acc(\%) & acc(\%) & acc(\%) & acc(\%) & acc(\%) & acc(\%) & acc(\%) \\
            \midrule
                        & \cellcolor{mygrey}-  &\cellcolor{mygrey} \textbf{97.3}  & \cellcolor{mygrey}\textbf{99.1}  & \cellcolor{mygrey}\textbf{84.4}  & \cellcolor{mygrey}\textbf{96.2}  &\cellcolor{mygrey}\textbf{87.4}  &\cellcolor{mygrey}\textbf{98.7}  &\cellcolor{mygrey}\textbf{70.5}  & \cellcolor{mygrey}\textbf{91.3}    \\
                        Convolution                      & TASN                                                                       & 97.9    & 99.3    &  89.1   &  98.6   &  89.3   & 98.9    &  87.9   &  97.9   \\
                        \&Attention            & ACNet                                           & 98.0    & 99.4    &  87.8   &  98.0   & 88.0    &  98.9   &  88.1   &  97.9   \\
                         Mechanisms            & FFVT                                                                       & 98.1    & 99.5    &  90.1   &  98.9   & 90.4    &  99.3   &  \textbf{91.8}   &  99.6   \\
                         (Our DCNN)                & RAMS-Trans               & \textbf{99.0}    & \textbf{99.9}    & \textbf{91.6}    & \textbf{99.8}    & \textbf{91.2}    & \textbf {99.6}   &  \textbf{91.8}   & \textbf{99.9}   \\
                                               & TransFG                                                                    & 98.3    & 99.6    & 90.4    & 99.1    & 89.6    &  99.0   &  90.0   &  99.5   \\
            \midrule
                        &\cellcolor{mygrey}-  &\cellcolor{mygrey}81.4  &\cellcolor{mygrey}95.6  &\cellcolor{mygrey}73.3  &\cellcolor{mygrey}89.3   &\cellcolor{mygrey}76.4   &\cellcolor{mygrey}91.4   &\cellcolor{mygrey}62.1  & \cellcolor{mygrey}78.3     \\
            Convolution                         & TASN~\cite{19}                                                                       & 88.1     & 98.7   & 81.5    & 95.6    & 86.3    & 97.5    & 85.6    & 95.9    \\
              Mechanism                                 & ACNet~\cite{20}                                                                      & 89.2     & 98.9   & 78.6    & 93.6    & 78.7    & 95.8    & 86.4    & 97.7    \\
            \midrule
            \multirow{4}{*}{}        
                        &\cellcolor{mygrey}-   &\cellcolor{mygrey}86.2   &\cellcolor{mygrey}97.9   &\cellcolor{mygrey}81.7    &\cellcolor{mygrey}95.6   &\cellcolor{mygrey}79.7   &\cellcolor{mygrey}96.1   &\cellcolor{mygrey}69.0  & \cellcolor{mygrey} 80.4   \\
                         Attention                    & FFVT~\cite{24}                                                                       & 94.1     & 99.0   & 88.0    & 98.4    & 89.0    & 99.0    & 90.0    &  99.0   \\
                        Mechanism                    & RAMS-Trans~\cite{25}                                                                 & 96.9     & 99.1   & 89.3    & 98.9    & 90.5    & 99.0    & 90.7    &  99.2    \\
                                               & TransFG~\cite{26}                                                                    & 95.4     & 99.0   & 88.9    & 98.6    & 88.8    & 98.9    & 89.9    &  98.0   \\
            \bottomrule
        \end{tabular}}
    \caption{Classification results of the proposed DCNN, convolution mechanism-based backbones, and attention mechanism-based backbones were combined with various Pre-Methods (preprocessing methods) on various fine-grained datasets. “-” means “without Pre-Methods”. All models did not use transfer learning methods to pre-train parameters. Convolution mechanism-based backbones with TASN, convolution mechanism-based backbones with ACNet, attention mechanism-based backbones with FFVT, attention mechanism-based backbones with RAMS-Trans, and attention mechanism-based backbones with TransFG were literature~\cite{19,20,24,25,26} used as the main backbone models with preprocessing methods. Adam optimized all models on the Oxford Flowers, Food101, Stanford Dogs, and CUB200-2021 data sets. The batch size is set to 128, the learning rate is 0.0001, and the training process is 300 epochs (Oxford Flowers), 200 epochs (Food101), 100 epochs (Stanford Dogs), and 100 epochs (CUB200-2021).}
    \label{table:4}
\end{table*}

\begin{figure*}[t]
    \centering
     \includegraphics[width=0.9\linewidth]{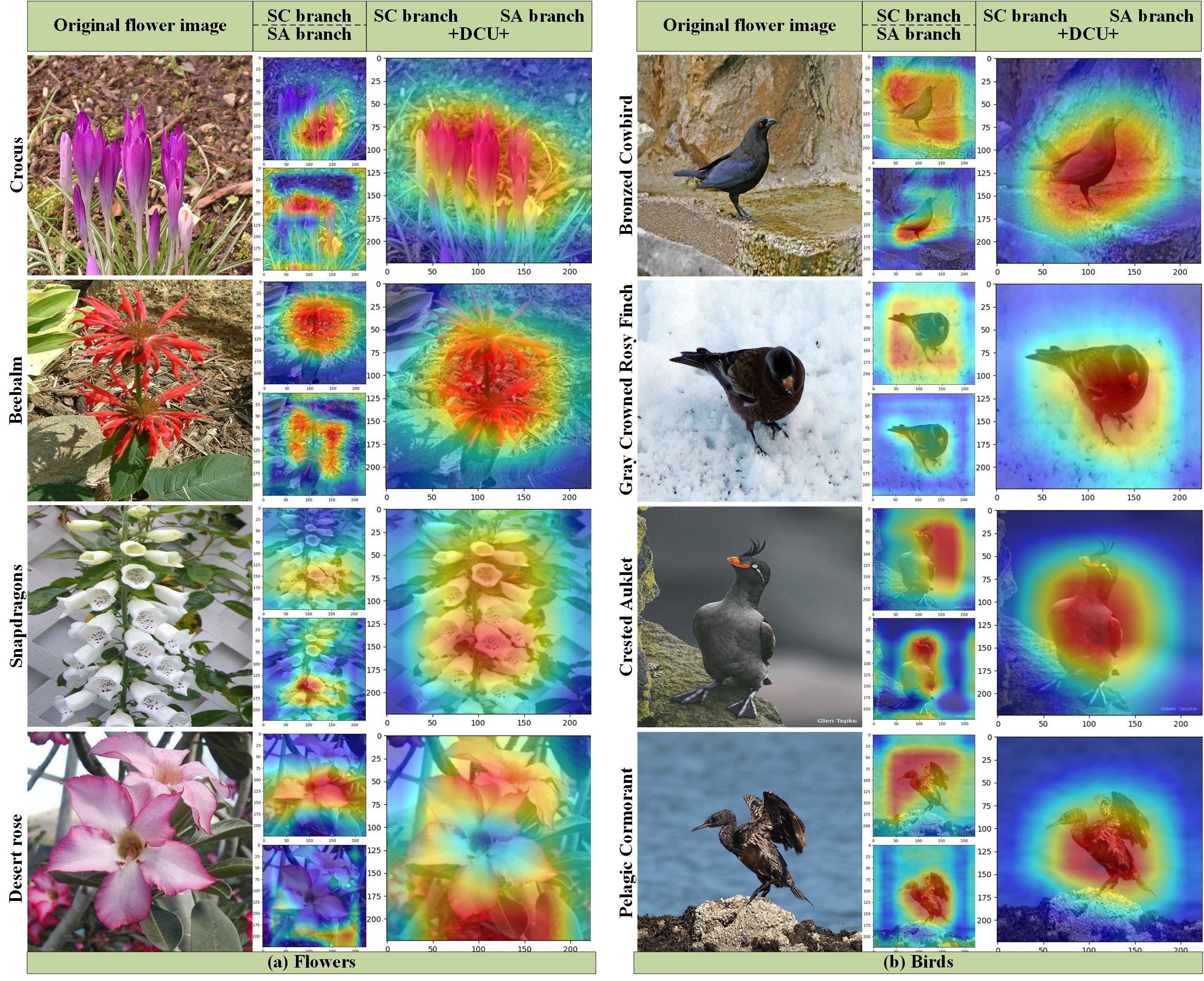}
     \caption{CAM visualization showing the effect through the learning process of the proposed DCNN, SC, and SA branches on the (a) Oxford Flowers and (b) CUB200-2021 dataset: comparative CAM image results between DCNN and SC/SA branches. DCNN fusions local features of SC branches and global representations of SA branches using DCU interactive bridges, better guiding attention to flowers and birds. Crocus, Beebalm, Snapdragons, and Desert rose are flower names; Pelagic Cormorant, Crested Auklet, Gray Crowned Rosy Finch, and Bronzed Cowbird are bird names. The red-colored parts are the features that the model focuses on during feature extraction.}
     \label{fig:5}
  \end{figure*}

\definecolor{mygrey}{RGB}{220,220,220}
\begin{table*}[H]
    \centering
    \resizebox{0.9\linewidth}{!}{
        \begin{tabular}{cccccccccc}
            \toprule
                                               & \multirow{3}{*}{\diagbox{\rule[-2mm]{0pt}{2mm}Pre-Methods}{\rule{0pt}{4mm}Datasets}} &
            \multicolumn{2}{c}{Oxford Flowers} &
            \multicolumn{2}{c}{Food101}        &
            \multicolumn{2}{c}{Stanford Dogs}  &
            \multicolumn{2}{c}{CUB200-2021}                                                                                                                                                                           \\
            \cmidrule(r){3-4}\cmidrule(lr){5-6}\cmidrule(lr){7-8}\cmidrule(l){9-10}
            Mechanisms                          &                                                                                      & Top1-   & Top5-   & Top1-   & Top5-   & Top1-   & Top5-   & Top1-   & Top5-   \\
                                               &                                                                                      & acc(\%) & acc(\%) & acc(\%) & acc(\%) & acc(\%) & acc(\%) & acc(\%) & acc(\%) \\
            \midrule
                        & \cellcolor{mygrey}-  &\cellcolor{mygrey} \textbf{97.3}  & \cellcolor{mygrey}\textbf{99.1}  & \cellcolor{mygrey}\textbf{84.4}  & \cellcolor{mygrey}\textbf{96.2}  &\cellcolor{mygrey}\textbf{87.4}  &\cellcolor{mygrey}\textbf{98.7}  &\cellcolor{mygrey}\textbf{70.5}  & \cellcolor{mygrey}\textbf{91.3}    \\
                        Convolution                      & TASN                                                                       & 97.9    & 99.3    &  89.1   &  98.6   &  89.3   & 98.9    &  87.9   &  97.9   \\
                        \&Attention            & ACNet                                           & 98.0    & 99.4    &  87.8   &  98.0   & 88.0    &  98.9   &  88.1   &  97.9   \\
                         Mechanisms            & FFVT                                                                       & 98.1    & 99.5    &  90.1   &  98.9   & 90.4    &  99.3   &  \textbf{91.8}   &  99.6   \\
                         (Our DCNN)                & RAMS-Trans               & \textbf{99.0}    & \textbf{99.9}    & \textbf{91.6}    & \textbf{99.8}    & \textbf{91.2}    & \textbf {99.6}   &  \textbf{91.8}   & \textbf{99.9}   \\
                                               & TransFG                                                                    & 98.3    & 99.6    & 90.4    & 99.1    & 89.6    &  99.0   &  90.0   &  99.5   \\
            \midrule
                        &\cellcolor{mygrey}-  &\cellcolor{mygrey}81.4  &\cellcolor{mygrey}95.6  &\cellcolor{mygrey}73.3  &\cellcolor{mygrey}89.3   &\cellcolor{mygrey}76.4   &\cellcolor{mygrey}91.4   &\cellcolor{mygrey}62.1  & \cellcolor{mygrey}78.3     \\
            Convolution                         & TASN~\cite{19}                                                                       & 88.1     & 98.7   & 81.5    & 95.6    & 86.3    & 97.5    & 85.6    & 95.9    \\
              Mechanism                                 & ACNet~\cite{20}                                                                      & 89.2     & 98.9   & 78.6    & 93.6    & 78.7    & 95.8    & 86.4    & 97.7    \\
            \midrule
            \multirow{4}{*}{}
                        &\cellcolor{mygrey}-   &\cellcolor{mygrey}86.2   &\cellcolor{mygrey}97.9   &\cellcolor{mygrey}81.7    &\cellcolor{mygrey}95.6   &\cellcolor{mygrey}79.7   &\cellcolor{mygrey}96.1   &\cellcolor{mygrey}69.0  & \cellcolor{mygrey} 80.4   \\
                         Attention                    & FFVT~\cite{24}                                                                       & 94.1     & 99.0   & 88.0    & 98.4    & 89.0    & 99.0    & 90.0    &  99.0   \\
                        Mechanism                    & RAMS-Trans~\cite{25}                                                                 & 96.9     & 99.1   & 89.3    & 98.9    & 90.5    & 99.0    & 90.7    &  99.2    \\
                                               & TransFG~\cite{26}                                                                    & 11     & 11   &11    & 11   & 11    & 11   & 11    &  11   \\
            \bottomrule
        \end{tabular}}
    \caption{Classification results of the proposed DCNN, convolution mechanism-based backbones, and attention mechanism-based backbones were combined with various Pre-Methods (preprocessing methods) on various fine-grained datasets. “-” means “without Pre-Methods”. All models did not use transfer learning methods to pre-train parameters. Convolution mechanism-based backbones with TASN, convolution mechanism-based backbones with ACNet, attention mechanism-based backbones with FFVT, attention mechanism-based backbones with RAMS-Trans, and attention mechanism-based backbones with TransFG were literature~\cite{19,20,24,25,26} used as the main backbone models with preprocessing methods. Adam optimized all models on the Oxford Flowers, Food101, Stanford Dogs, and CUB200-2021 data sets. The batch size is set to 128, the learning rate is 0.0001, and the training process is 300 epochs (Oxford Flowers), 200 epochs (Food101), 100 epochs (Stanford Dogs), and 100 epochs (CUB200-2021).}
    \label{table:4}
\end{table*}

\cref{table:4} presents this at a certain level of computing power. DCNN performance compared to advanced models on five fine-grained datasets, where the proposed DCNN backbone network achieves higher top-1 and top-5 accuracy than ResNet-50 and ViT-B-16, and combined with the preprocessing method RAMS-Trans achieves the highest classification accuracy. Demonstrates the advantages of our proposed model as a fine-grained classification backbone network.
\cref{fig:5} (a) Flowers and (b) Birds show visualizations of feature extraction during DCNN training on the Oxford Flowers and CUB200-2011 datasets, respectively. The DCNN includes the SC branch, DCU, and SA branch. Among them, the convolutional algorithm in the SC branch is more inclined to focus on the local features of the flower and bird. In contrast, the attention mechanism in the SA branch is more inclined to focus on the global representation of the flower and bird. The two branches continuously and interactively fuse information through the DCU, guiding the attention to ignore background interference and the attraction of local features so as to accurately capture the overall information of flowers and birds, thereby improving classification performance.

\subsection*{Ablation Studies}

In this study, ablation experiments of the DCNN were performed in the ImageNet-1k dataset. \cref{table:5} shows the performance comparison of DCNN using different methods. With comparable performance, the separable convolution method in the SC branch has the advantage of fewer parameters than the conventional convolution method. With a comparable number of parameters, DCNN using the average pooling (Avg pooling) method in the DCU module can achieve higher performance. In the SA branch, the number of self-attention heads reaches 16. With a comparable number of parameters, better Top1-acc is obtained, verifying that the increase in the scale of this model will gain the advantage of better performance.

\begin{table}[H]
    \centering
        \begin{tabular}{cccc}
          \hline
           Branch & Methods & \#Params & Top1-acc (\%) \\ \hline
           \multirow{2}{*}{\begin{tabular}[c]{@{}c@{}}SC\\ branch\end{tabular}} & \begin{tabular}[c]{@{}c@{}}Conventional\\ Convolution\end{tabular} & 87.3M & 82.2 \\ \cline{2-4}
           & \begin{tabular}[c]{@{}c@{}}Separable\\ Convolution\end{tabular} & 52.7M & 82.3 \\ \hline
           \multirow{3}{*}{DCU} & \begin{tabular}[c]{@{}c@{}}Down: Convolution\\ Up: Interpolation\end{tabular} & 65.4M & 82.3 \\ \cline{2-4}
           & \begin{tabular}[c]{@{}c@{}}Down: Max pooling\\ Up: Interpolation\end{tabular} & 52.7M & 11 \\ \cline{2-4}
           & \begin{tabular}[c]{@{}c@{}}Down; Avg pooling\\ Up: Interpolation\end{tabular} & 11M & 11\\ \hline
           \multirow{2}{*}{\begin{tabular}[c]{@{}c@{}}SA\\ branch\end{tabular}} & \#Heads: 12 &11M & 11 \\ \cline{2-4}
           & \#Heads: 16 & \textbf{11M} & \textbf{11} \\ \hline
       \end{tabular}
    \caption{Comparison of DCNN performance (number of Params and Top1-acc) using different methods in the SC branch, DCU, and SA branch. Following the single-variable principle, the DCNN model obtains the performance by changing one method in each experiment.}
    \label{table:5}
\end{table}

\begin{table}[H]
    \centering
        \begin{tabular}{cccc}
        \hline
         Combined Models & DCU & \#Params & Top1-acc (\%) \\ \hline
         ResNet-101\&DeiT-S & × & 66.5M & 81.4 \\
         DCNN & × & 11M & 81.6 \\
         ResNet-101\&DeiT-S & \checkmark & \textbf{68.8M} & 82.0 \\
         DCNN & \checkmark & 11M & \textbf{82.3} \\ \hline
       \end{tabular}
    \caption{Comparison of the performance (number of Params and Top1-acc) of the combined models. “×” indicates that the combined model does not use the DCU module, and “\checkmark” indicates that the combined model uses the DCU module.}
    \label{table:6}
\end{table}

As one of the original core designs of DCNN, the proposed DCU module plays an enhanced bridging role in the interactive learning of SC and SA branches. The ablation experimental results in Table 6 reflect the influence of the DCU module on the performance of the combined models. For a comparable number of parameters, the combined model with the aid of DCU module achieves higher performance. Our proposed DCNN designed with the DCU module achieved the best performance.

\subsection*{Experimental analysis in brain tumor MRI dataset}

\begin{figure}[htb]
    \centering
     \includegraphics[width=0.5\linewidth]{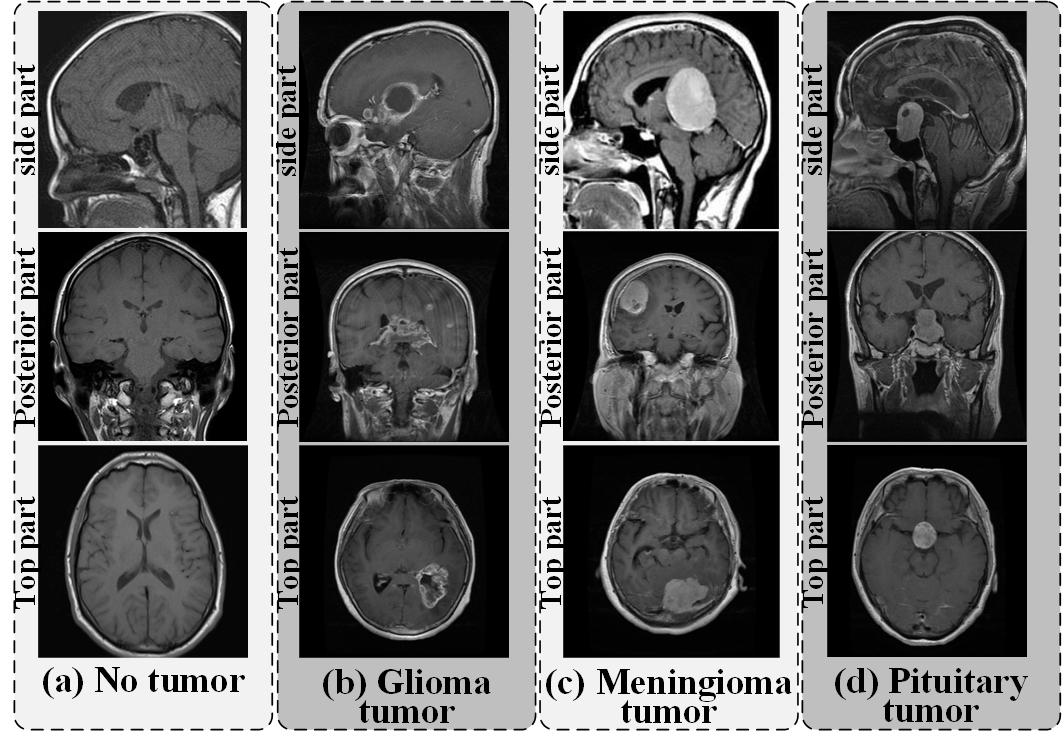}
     \caption{Four categories of brain MRI samples: (a) No tumor; (b) Glioma tumor; (c) Meningioma tumor; (d) Pituitary tumor. Each category covers three views of brain structures, including the top part, posterior part, and side part samples. All brain tumor MRI images come from public open databases without any privacy information.}
     \label{fig:6}
  \end{figure}

Next, 10,183 brain MRI samples were obtained by collecting online images and merging existing public databases (\url{https://www.kaggle.com/}). The samples were of four types: non-tumors (2,396), gliomas (2,547), meningiomas (2,582), and pituitary tumors (2,658), with a relatively balanced range across categories. The size of the dataset used in this experiment is considered above average for this field of study. The training and testing sets were divided in a 9:1 ratio. \cref{fig:6} shows four categories in the brain tumor MRI dataset, including three tumor categories and no tumor: (a) No tumor, (b) Glioma tumor, (c) Meningioma tumor, and (d) Pituitary tumor.

\definecolor{mygrey}{RGB}{220,220,220}
\begin{table}[htb]
    \centering
    \resizebox{0.4\linewidth}{!}{
        \begin{tabular}{ccc}
            \toprule
            Mechanisms                         & Pre-Methods     & Accuracy (\%) \\
            \midrule
                                              & \cellcolor{mygrey}-               & \cellcolor{mygrey}87.3          \\
            \cmidrule{2-3}
            Convolution                        & TASN~\cite{19}       & 91.2          \\
             Mechanism                                 & ACNet~\cite{20}      & 90.3          \\
            \midrule
            \multirow{4}{*}{}      & \cellcolor{mygrey}-               & \cellcolor{mygrey}88.1          \\
            \cmidrule{2-3}
              Attention                                & FFVT~\cite{24}       & 91.9          \\
             Mechanism                               & RAMS-Trans~\cite{25} & 93.5          \\
                                              & TransFG~\cite{26}    & 92.7          \\
            \midrule
            \multirow{6}{*}{} & \cellcolor{mygrey}-               & \cellcolor{mygrey}\textbf{90.7}          \\
            \cmidrule{2-3}
              Convolution                                & TASN       & 93.3          \\
              \&Attention                                & ACNet      & 92.8          \\
            Mechanisms                           & FFVT      & 94.1          \\
               (Our DCNN)                               & RAMS-Trans & \textbf{95.0}          \\
                                              & TransFG    & 93.9          \\
            \bottomrule
        \end{tabular}}
    \caption{Classification results of convolution mechanism-based backbones, attention mechanism-based backbones, and the proposed DCNN backbone networks were combined with various Pre-Methods (preprocessing methods) on the brain tumor MRI dataset. “-” indicates “without Pre-Methods”. TASN, ACNet, FFVT, RAMS-Trans, and TransFG were literature~\cite{19,20,24,25,26} used as the preprocessing methods. Adam optimized all models on the brain tumor MRI dataset. The batch size is 128, the learning rate is 0.0001, and the training process is 150 epochs.}
    \label{table:7}
    
\end{table}

\cref{table:7} presents the accuracy of various models from various studies (see the references for detailed information). The DCNN demonstrates the best classification ability among all the mentioned backbone models. In other words, the proposed DCNN model uses the RAMS-Trans preprocessing method, outperforms the advanced models, and achieves better classification and recognition results than the existing models.

\section*{Conclusions}

In this study, we propose the DCNN backbone network to improve the classification performance of fine-grained images. By designing the SC branch, DCU bridge module, and SA branch, we constructed novel learning networks that incorporate the advantages of convolutional operators and attention mechanisms that can extract local features and global features of fine-grained images. In addition, we enhanced the interactive learning of the two types of features by superposing the bridge information for better recognition of fine-grained features. We performed applied experiments in the brain tumor MRI dataset. The experiment results expressed that the proposed DCNN combined with the preprocessing method RAMS-Trans achieved the best classification accuracy compared to other advanced fine-grained image classification models.

For future works, the proposed DCNN can be developed as an advanced fine-grained image classifier suitable for medical imaging (a typical fine-grained image recognition task) by applying transfer learning, reinforcement learning, and integrated learning mechanisms. Thus, our approach can contribute to promoting and using artificial intelligence technology in clinical medical imaging diagnosis.

\bibliography{main}

\noindent LaTeX formats citations and references automatically using the bibliography records in your .bib file, which you can edit via the project menu. Use the cite command for an inline citation, e.g.  \cite{Hao:gidmaps:2014}.

For data citations of datasets uploaded to e.g. \emph{figshare}, please use the \verb|howpublished| option in the bib entry to specify the platform and the link, as in the \verb|Hao:gidmaps:2014| example in the sample bibliography file.

\section*{Acknowledgements}

This work was supported by the Shandong Province Outstanding Youth Science Fund 
Project (Overseas) (2023HWYQ-098). 

\section*{Author contributions statement}

Must include all authors, identified by initials, for example:
A.A. conceived the experiment(s),  A.A. and B.A. conducted the experiment(s), C.A. and D.A. analysed the results.  All authors reviewed the manuscript.





\end{document}